\documentclass[]{spie}  

\usepackage{amsmath,amsfonts,amssymb}
\usepackage{graphicx}
\usepackage{booktabs}
\usepackage[colorlinks=true, allcolors=blue]{hyperref}

\title{Compressibility Analysis for the differentiable shift-variant Filtered Backprojection Model}

\author[a]{Chengze~Ye}
\author[a,b]{Linda-Sophie~Schneider}
\author[a,b]{Yipeng~Sun}
\author[a]{Mareike~Thies}
\author[a]{Andreas~Maier}
\affil[a]{Friedrich-Alexander University Erlangen-Nuremberg, Erlangen, Germany}
\affil[b]{Fraunhofer EZRT, Flugplatzstraße 75, 90768 Fürth, Germany}

\authorinfo{Further author information: (Send correspondence to Chengze~Ye)\\Chengze~Ye: E-mail: chengze.ye@fau.de, Telephone: +49 9131 85-27775}

\begin{document} 
\maketitle

\begin{abstract}

The differentiable shift-variant filtered backprojection (FBP) model enables the reconstruction of cone-beam computed tomography (CBCT) data for any non-circular trajectories. This method employs deep learning technique to estimate the redundancy weights required for reconstruction, given knowledge of the specific trajectory at optimization time. However, computing the redundancy weight for each projection remains computationally intensive.
This paper presents a novel approach to compress and optimize the differentiable shift-variant FBP model based on Principal Component Analysis (PCA). 
We apply PCA to the redundancy weights learned from sinusoidal trajectory projection data, revealing significant parameter redundancy in the original model. By integrating PCA directly into the differentiable shift-variant FBP reconstruction pipeline, we develop a method that decomposes the redundancy weight layer parameters into a trainable eigenvector matrix, compressed weights, and a mean vector. This innovative technique achieves a remarkable $97.25\%$ reduction in trainable parameters without compromising reconstruction accuracy. 
As a result, our algorithm significantly decreases the complexity of the differentiable shift-variant FBP model and greatly improves training speed. These improvements make the model substantially more practical for real-world applications.



\end{abstract}


\section{INTRODUCTION}
\label{sec:intro}  
Cone beam computed tomography (CBCT) is a commonly utilized imaging technology in the field of interventional medicine. As the name implies, the technology employs a cone-shaped X-ray beam that covers a substantial volume in a single rotation, enabling the acquisition of three-dimensional data with less radiation exposure. 

In recent years, robotic C-arm imaging devices have been developed with highly flexible movement capabilities, enabling scans along various non-circular trajectories. This not only accelerates the process of data acquisition but also effectively mitigates the impact of metal artifacts \cite{herl2020scanning}. It is therefore imperative to develop a rapid and precise reconstruction algorithm for non-circular trajectory CBCT.

Iterative reconstruction algorithms are commonly employed for the reconstruction of CBCT data acquired along non-circular trajectories. These algorithms continuously adjust the image at each iteration in order to gradually approach the actual data, resulting in reconstruction speeds that are relatively slow and high resource consumption. 

To address this issue, Defrise and Clack \cite{defrise1994cone} proposed a shift-variant filtered backprojection (FBP) algorithm that significantly improves the reconstruction speed for non-circular trajectory CBCT. However, for each type of trajectory geometry, it is necessary to recalculate the redundancy weights in the filter. In the case of complex non-circular trajectory geometries, the challenge of solving the corresponding redundancy weights analytically is considerable.

In our previous study, We built on the shift-variant FBP algorithm and employed the known operator learning technique to construct a differentiable shift-variant FBP model \cite{ye2024deep, ye2024draco}. This data-driven approach enables the automatic estimation of appropriate redundancy weights based on the trajectory geometry through the training process. While this method effectively estimates redundancy weights, estimating a redundancy weight for each projection still requires a substantial computational burden. 

This paper aims to address the aforementioned challenge by incorporating PCA into the reconstruction pipeline of the differentiable shift-variant FBP model. 
By representing high-dimensional redundancy weight parameters with compressed low-dimensional redundancy weights and the corresponding linear transformation, a significant reduction in redundant parameters is achieved while maintaining reconstruction accuracy. This approach markedly accelerates the training process and enhances the practical applicability of the algorithm.

\section{Methods}
\subsection{Shift-Variant FBP Algorithm}
\label{sec:2.1}

Defrise and Clack \cite{defrise1994cone} propose a shift-variant FBP algorithm employed for the reconstruction of CBCT data with any specific trajectories. The algorithm permits the simultaneous execution of projection acquisition and processing, thereby markedly accelerating the reconstruction process.
The reconstruction formula is as follows:
\begin{align*}
f(x)=&\int_{ \Lambda}^{} d\lambda\int_{S^2/2}^{}d\theta-\frac{1}{4\pi^2}\mid a'(\lambda)\cdot\theta\mid\frac{1}{n(\theta, \lambda)}\\
&\times\delta' ((x-a(\lambda))\cdot \theta)S(\theta, \lambda ).\tag{1}
\label{eq:myequation3}
\end{align*}

The term $n(\theta, \lambda)$ denotes the number of intersections between trajectory $a(\Lambda)$ and the plane orthogonal to $\theta$ passing through the point $a(\lambda)$ on the trajectory. The term $\frac{1}{n(\theta, \lambda)}$ and $\mid a'(\lambda)\cdot\theta\mid$  are intricately linked to the trajectory's geometry and could be regarded as a redundancy weight. 
In other words, the redundancy weights need to be calculated for every new trajectory.

\subsection{Differentiable Shift-Variant FBP Model}
\label{sec:2.2.1}

The differentiable shift-variant FBP model \cite{ye2024deep, ye2024draco} represents a innovative data-driven reconstruction pipeline that builds upon the traditional shift-variant FBP algorithm.

This approach transforms the original algorithm into a differentiable neural network architecture by applying known operator learning. A key feature of this network is its use of backpropagation to estimate redundancy weights based on given trajectory geometries, enhancing its adaptability and accuracy.

The conversion of Eq.~\eqref{eq:myequation3} into a neural network representation yields Eq.~\eqref{eq:myequation15}.
\begin{align*}
x=A_{3d}^Tw_{d}A_{2d}^TDw_{red}w_{sino}DA_{2d}w_{cos}p\tag{2}
\label{eq:myequation15}
\end{align*}
The reconstruction pipeline includes a series of operations on cone beam projections $p$, such as the application of different weights $w_{cos}$, $w_{sino}$, $w_{red}$, $w_{d}$, differentiation $D$, the 2D Radon transform $A_{2d}$, the 2D backprojection $A_{2d}^T$, and finally, the reconstruction of the volume through 3D backprojection $A_{3d}^T$.

\subsection{PCA-Based Network Compression}

PCA \cite{mackiewicz1993principal} is a widely utilized technique for reducing the dimensionality of data. This technique reduces the dimensionality and complexity of high-dimensional data by projecting it into a lower-dimensional space while preserving as much of the original information as possible. 

This chapter will examine the application of this method to the compression of redundancy weights in the reconstruction pipeline, thereby achieving a reduction in the number of parameters.

PCA is comprised of five steps:

\begin{enumerate}
\item Standardize the Data: If the features in the data have different scales, standardize the data so that each feature has a mean of 0 and a standard deviation of 1. This ensures that each feature contributes equally to the analysis.
\begin{align*}
\tilde{X} = X - \mu
\tag{3}
\label{eq:myequation5}
\end{align*}
\item Compute the Covariance Matrix: Calculate the covariance matrix to understand the linear relationships between the features.
\begin{align*}
\Sigma = \frac{1}{n-1} \tilde{X}^T \tilde{X}
\tag{4}
\label{eq:myequation8}
\end{align*}

\item Compute the Eigenvalues and Eigenvectors: Perform eigenvalue decomposition on the covariance matrix to obtain the eigenvalues and corresponding eigenvectors. 
\item Sort the eigenvalues in descending order and select the top k eigenvalues. The corresponding eigenvectors are the principal components $V_k = [v_1, v_2, \ldots, v_k]$.
\item Transform the Data: Project the original data onto the selected principal components to obtain the reduced-dimension data set.
\begin{align*}
Y = \tilde{X} V_k
\tag{5}
\label{eq:myequation7}
\end{align*}
\end{enumerate}

Project the dimensionally reduced data Y back to the higher-dimensional space, then add the original mean to the centered data to obtain the reconstructed data matrix $X'$.
\begin{align*}
X' = Y V_k^T + \mu
\tag{6}
\label{eq:myequation10000}
\end{align*}

\begin{figure}[!t]
\centering
\includegraphics[width=2.2in]{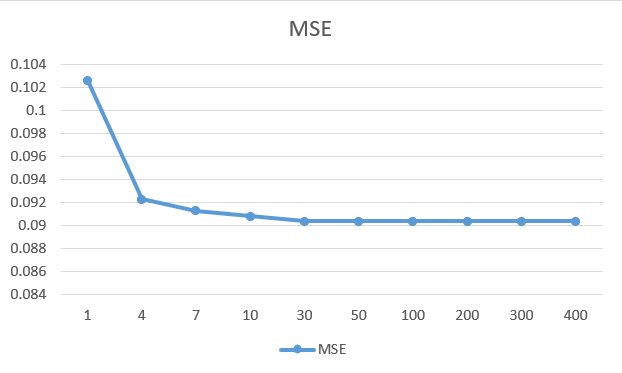}
\includegraphics[width=2.2in]{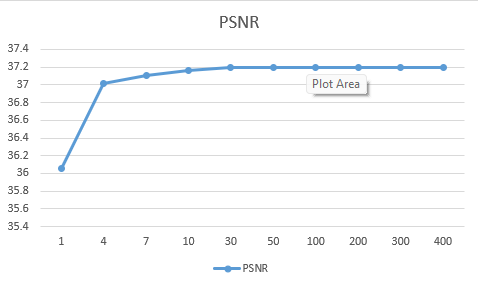}
\includegraphics[width=2.2in]{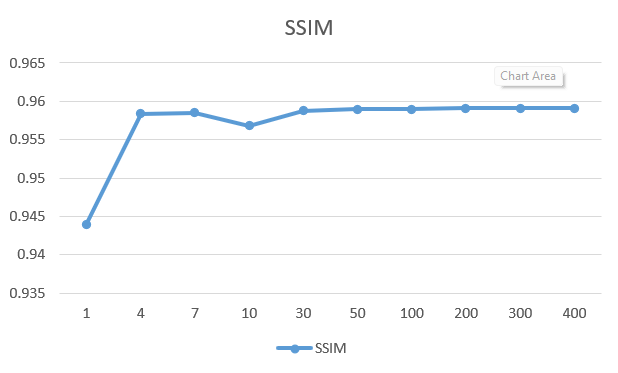}
\caption{Reconstruction performance of redundancy weights using different numbers of principal components.}
\label{ssim}
\end{figure}

Fig. \ref{ssim} presents the analysis of the reconstruction performance of redundancy weights, restored using different numbers of principal components, evaluated through MSE, SSIM, and PSNR metrics.
It can be observed that even when the number of principal components used for dimensionality reduction and reconstruction is gradually reduced to only 30, the impact on reconstruction quality is minimal, while the dimensionality of redundancy weights is significantly reduced. The application of this concept to the training of neural networks can also result in a notable reduction in the number of trainable parameters.

In this paradigm, neural networks initially learn to represent redundancy weights in a low-dimensional manifold ($w_{red}'$), followed by the acquisition of principal components ($V_k$) that serve as a sophisticated mapping function to the original high-dimensional space. The process culminates in the reintegration of the estimated mean ($\mu$) into the centered redundancy weights, thereby achieving a remarkable reduction in trainable parameters without sacrificing model expressivity. This represents a significant advancement in the ongoing quest for more efficient and powerful machine learning paradigms.

The integration of this idea into the differentiable Shift-Variant FBP model yields the following formula:
\begin{align*}
x=A_{3d}^Tw_{d}A_{2d}^TD(w_{red}' \times V_k^T + \mu)w_{sino}DA_{2d}w_{cos}p.\tag{7}
\label{eq:myequation9}
\end{align*}

\begin{figure*}[!t]
\centering
\includegraphics[width=6.8in]{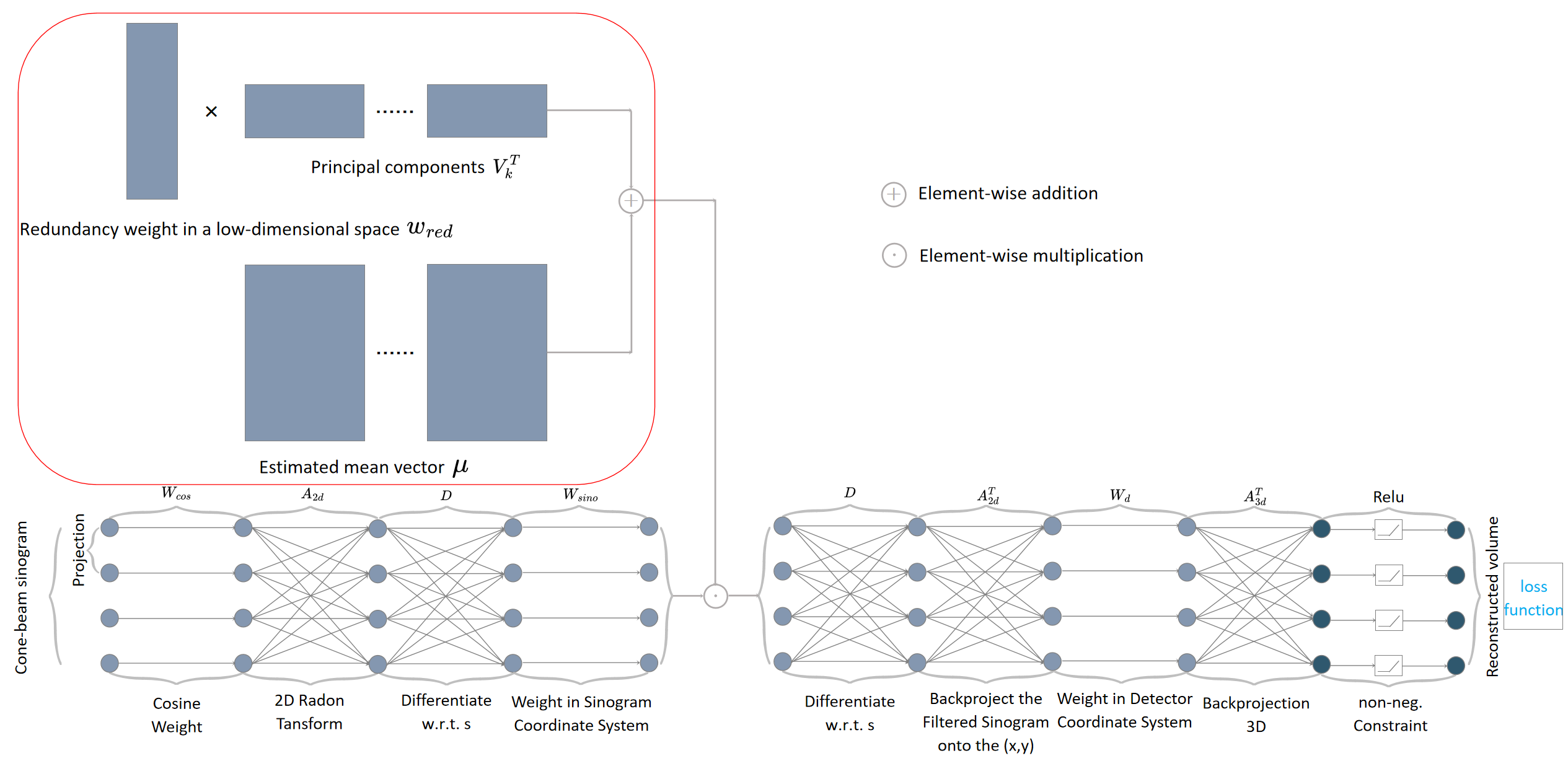}
\caption{Compressed differentiable shift-variant FBP model: Low-dimensional redundancy weights are initially mapped back to the high-dimensional space, and subsequently employed to replace the redundancy weights within the differentiable Shift-Variant FBP model.}
\label{Architecture}
\end{figure*}

Based on Eq.~\eqref{eq:myequation9}, the complete network architecture has been constructed, as shown in Fig.~\ref{Architecture}. 


\section{Experiments}

The loss function for training is formulated as:
\begin{align*}
\mathcal{L}_{total} = \mathcal{L}_{mse} + \lambda \cdot\mathcal{L}_{ssim},\tag{8}
\label{eq:myequation4}
\end{align*}
where $\mathcal{L}_{mse}$ represents the mean squared error (MSE) loss, $\mathcal{L}_{ssim}$ denotes structural similarity index (SSIM) loss, and $\lambda$ is a weighting factor. The application of the One-cycle learning rate policy \cite{smith2018disciplined}, which enables the adjustment of the learning rate within a specified range, resulted in a notable enhancement in the training performance. The training process was conducted on an Nvidia $A40$ GPU.

In our experiments, we employed the precise geometry parameters of a clinical C-arm system, the Artis zeego (Siemens AG, Forchheim, Germany). The positions of the sources follow the sinusoidal trajectory.

In accordance with the aforementioned trajectory geometry, a total of $30$ simulated data samples were generated using the cone beam forward projection in the PyroNN package \cite{syben2019pyro} in order to create a suitable training and validation dataset. $24$ samples were utilized for training, while the remaining samples were employed for validation. 

In order to assess the efficacy of our method in the domain of medical imaging, the Pancreatic-CT-CBCT-SEG dataset \cite{hong2021breath} was employed. Subsequently, the corresponding sinogram was generated the same as the simulated data generation.

\section{Results}

\begin{figure}[!t]
\centering
\includegraphics[width=5.5in]{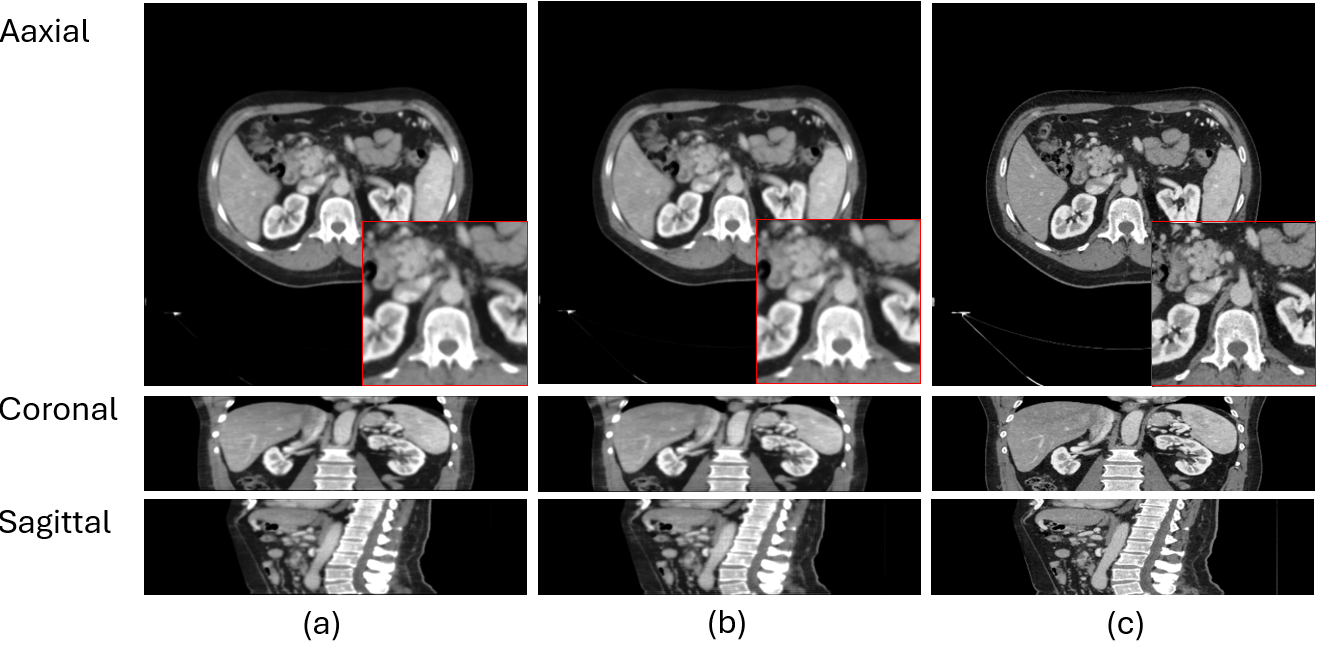}
\caption{Reconstruction result with PCA-based dimensionality reduction. (a) Reconstruction without PCA-based dimensionality reduction. (b)Reconstruction with PCA-based dimensionality reduction (10 Components). (c)Ground truth as reference.}
\label{result(PCA)}
\end{figure}
\begin{figure}[!t]
\centering
\includegraphics[width=4in]{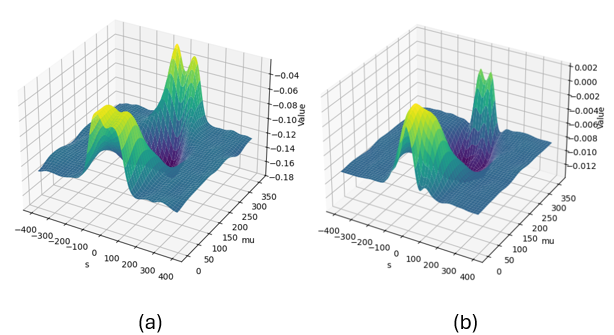}
\caption{Redundancy weights comparison. (a) Redundancy weights recovered from the low-dimensional representation. (b) Learned redundancy weights without compression.}
\label{redundancy_comparison}
\end{figure}

The neural network achieved convergence after $140$ epochs, whereas in the case of the uncompressed network, this number was $430$, thereby significantly accelerating the training process. 
Fig.~\ref{result(PCA)} depicts the outcomes of the neural network following convergence. It can be observed that even with a significant reduction in parameters, high reconstruction quality can still be achieved. Moreover, as demonstrated in Figure~\ref{redundancy_comparison}, the figure presents a comparison between the redundancy weights recovered from the low-dimensional representation and the learned redundancy weights without dimensionality reduction. The results of this comparison demonstrate a remarkable similarity, thereby further validating the effectiveness of the dimensionality reduction in reconstruction process.

Table \ref{tab:comparison} presents a quantitative analysis of the impact of reducing parameters on the reconstruction quality of the differentiable Shift-Variant FBP model, employing MSE, peak signal-to-noise ratio (PSNR), and SSIM as metrics. Additionally, the number in parentheses represents the number of retained principal components. The empirical findings demonstrate that the implementation of this parameter-reduction technique engenders a negligible compromise in reconstruction fidelity, despite effecting a substantial diminution in the neural network's parametric complexity. 

\begin{table}[htbp]
\centering
\caption{Comparison of Image Quality Metrics}
\begin{tabular}{@{}lcccccc@{}}
\toprule
& \textbf{MSE}$\downarrow$ & \textbf{PSNR (dB)}$\uparrow$ & \textbf{SSIM}$\uparrow$ & \textbf{Number of parameters}$\downarrow$ \\ 
\midrule
No PCA & 0.0904 ± 0.0149&  37.20± 1.34& 0.9591±0.0051 & 113,184,000\\
PCA(50) & 0.0867± 0.0130&  37.54± 1.34& 0.9427±0.0088 & 14,450,960\\ 
PCA(30) & 0.0876± 0.0132&  37.45± 1.34& 0.9409±0.0088 & 8,783,760\\ 
PCA(10) & 0.0855± 0.0131&  37.67± 1.34& 0.9481±0.0074 & 3,116,560\\ 
\bottomrule
\end{tabular}%
\label{tab:comparison}
\end{table}


\section{Conclusion}
This paper presents a novel integration of PCA into the differentiable Shift-Variant FBP model, yielding a remarkable $97.25\%$ reduction in trainable parameters while preserving reconstruction fidelity. This approach results in a notable reduction in model complexity, which in turn leads to a considerable acceleration in the training process and a substantial enhancement in the algorithm's practical viability. By examining the redundancy weights within the reconstruction pipeline, we have demonstrated that these parameters are compressible, thereby establishing a robust foundation for the development of efficient reconstruction methodologies applicable to arbitrary trajectories. Nevertheless, further investigation is required to examine the compression performance and rates of this method on more complex trajectories. In conclusion, this study makes a significant contribution to the practical application of the differentiable Shift-Variant FBP model.

\acknowledgments 
 
The research leading to these results has received funding from the Pattern Recognition Lab of the FAU Erlangen-Nürnberg.
The authors gratefully acknowledge the scientific support and HPC resources provided by the Erlangen National High Performance Computing Center (NHR@FAU) of the FAU Erlangen-Nürnberg.


\bibliographystyle{spiebib} 

\end{document}